\pdfoutput=1
\documentclass[11pt]{article}

\usepackage[]{acl}

\usepackage{times}
\usepackage{latexsym}

\usepackage[T1]{fontenc}
\usepackage[utf8]{inputenc}

\usepackage{microtype}

\usepackage{color}
\definecolor{ForestGreen}{RGB}{34,139,34}
\definecolor{Amber}{RGB}{255,191,0}

\usepackage{hyperref}

\usepackage[inline]{enumitem}
\usepackage{scrextend}
\usepackage{array}
\usepackage{arydshln}
\usepackage{graphicx}
\usepackage{hhline}

\usepackage{enumitem}

\usepackage{censor}
\usepackage{gb4e}
\noautomath

\usepackage{amsmath}

\usepackage{multirow}
\usepackage{makecell}

\usepackage{arydshln}
\usepackage{array}
\newcolumntype{L}[1]{>{\raggedright\let\newline\\\arraybackslash\hspace{0pt}}m{#1}}
\newcolumntype{C}[1]{>{\centering\let\newline\\\arraybackslash\hspace{0pt}}m{#1}}
\newcolumntype{R}[1]{>{\raggedleft\let\newline\\\arraybackslash\hspace{0pt}}m{#1}}

\usepackage{scrextend}

\title{Abstraction not Memory: BERT and the English Article System}

\author{
Harish Tayyar Madabushi\textsuperscript{1,2},
\\
\textbf{Dagmar Divjak\textsuperscript{3,4}} \and
\textbf{Petar Milin\textsuperscript{3}}
\\[0.3cm]
\textsuperscript{1} Department of Computer Science, University of Sheffield \\
\textsuperscript{2} School of Computer Science, University of Birmingham \\
\textsuperscript{3} Department of Modern Languages, University of Birmingham \\
\textsuperscript{4} Department of English Language and Linguistics, University of Birmingham \\
\texttt{\small H.TayyarMadabushi@sheffield.ac.uk} \\[-0.1mm]
\texttt{\small(D.Divjak, P.Milin)@bham.ac.uk}\\
\\
}

\begin{document}
\maketitle
\begin{abstract}
Article prediction is a task that has long defied accurate linguistic description. As such, this task is ideally suited to evaluate models on their ability to emulate native-speaker \emph{intuition}. To this end, we compare the performance of native English speakers and pre-trained models on the task of article prediction set up as a three way choice (a/an, the, zero). Our experiments with BERT show that BERT outperforms humans on this task across all articles. In particular, BERT is far superior to humans at detecting the zero article, possibly because we insert them using rules that the deep neural model can easily pick up. More interestingly, we find that BERT tends to agree more with annotators than with the corpus when inter-annotator agreement is high but switches to agreeing more with the corpus as inter-annotator agreement drops. We contend that this alignment with annotators, despite being trained on the corpus, suggests that BERT is not memorising article use, but captures a high level generalisation of article use akin to human intuition. 

\end{abstract}

\section{Introduction and Motivation}
\label{section:intro}
Pre-trained models, such as BERT~\citep{DBLP:journals/corr/abs-1810-04805}, RoBERTa~\citep{DBLP:journals/corr/abs-1907-11692} and more recently T5~\citep{JMLR:v21:20-074}, are the state of the art across several tasks in computational linguistics. In addition, transformer-based models are known to have access to information as varied as part of speech information~\citep{chrupala-alishahi-2019-correlating,tenney2018what}, parse trees~\cite{hewitt-manning-2019-structural}, the NLP pipeline~\cite{DBLP:journals/corr/abs-1905-05950}, and constructional information~\citep{tayyar-madabushi-etal-2020-cxgbert}. These models tend to perform so well that, on certain tasks, they outperform human baselines~\citep{zhang2020retrospective}. 

In this work, we investigate how well language models, specifically BERT Large, perform on the linguistically interesting task of article prediction. English article prediction, further discussed in Section \ref{section:eng-article-system}, is a phenomenon that native speakers of English find almost trivial. At the same time, linguists find it particularly difficult to formulate the rules that would govern article usage: article use cannot be captured by local co-occurrence but is dependent on the wider context  and often there is no one ``right'' article, but multiple options are possible, albeit with slight differences in the meaning conveyed. Grammar correction systems prior to BERT struggled to reach acceptable levels of performance on article selection (detailed in \mbox{Section \ref{section:eng-article-system}}). As we will show, BERT shows performance on this task that is superior to that of humans. Given this, it is interesting to investigate how BERT attains this level of accuracy and what the implications are for the system: does BERT manage to go beyond the local vicinity into the larger context to track the referent?   

The current study compares the performance of transformer-based pre-trained models and humans in an attempt to explore how language models handle an, in essence, \emph{creative} task, with an emphasis on how model performance changes with inter-annotator agreement. We also explicitly incorporate the plural indefinite or zero (Ø) article (detailed in \mbox{Section \ref{section:eng-article-system}}) as in the sentence \textit{There are Ø merchant bankers who find it convenient to stir up Ø apprehension with a view to drumming up Ø business for themselves.}

The flexibility that is inherent in article usage requires us to explore methods of evaluation that do not rely solely on accuracy. While the shortcomings of relying too heavily on accuracy based metrics have been highlighted in prior work (see Section \ref{section:related-work}), these difficulties are accentuated by the presence of flexibility. Clearly, there is little need to require a model to output one specific class if people are comfortable with multiple options. As such, we evaluate performance based on the Matthews correlation coefficient between human annotators and model outputs \emph{at each different level of inter-annotator agreement}. 

To this end, this works aims to answer the following questions: 
\begin{itemize*}
    \item[a)] How well do language models perform on a task that humans rely on intuition rather than deliberate reasoning, specifically article prediction, and
    \item[b)] how does this performance vary with increased flexibility in the article that can be used, as measured by inter-annotator agreement
\end{itemize*}. So as to ensure reproducibility and to aid future research in this direction, we make our scripts freely available and our dataset, built from the British National Corpus (BNC)~\cite{bnc2007british}, available under the required licence\footnote{\href{https://github.com/H-TayyarMadabushi/Abstraction-not-Memory-BERT-and-the-English-Article-System-NAACL-2022}{https://github.com/H-TayyarMadabushi/Abstraction-not-Memory-BERT-and-the-English-Article-System-NAACL-2022}}. Further details on the BNC are presented in Appendix \ref{app:bnc-licence}.

\section{The English Article System}  
\label{section:eng-article-system}

There are three articles in English: 
\begin{enumerate*}
    \item[a)] the definite article, \textit{the}, 
    \item[b)] the indefinite article, \textit{a/an}, and 
    \item[c)] the absence of an article or the \textit{zero (Ø) article}~\citep{swan1997english}.
\end{enumerate*}

%An early attempt to describe the English article system was by \citet{thomas_1989}, who proposed a simple system with four types of articles, dictated by two variables: Specific Referent, which signifies whether the noun the article is referring to is specific; and Hearer Knowledge, which signifies whether the hearer is aware of the noun being referred to. %This resulted in four types of articles with the addition of a fifth type to account for idiomatic instance of articles. 

There have been several sets of guidelines for the use of articles starting with the early works by \citet{huebner1983longitudinal,https://doi.org/10.1111/j.1467-1770.1985.tb01022.x,thomas_1989}. The most general ones rely on a few parameters only, such as Hearer Knowledge (whether the interlocutor can be considered to be able to identify the referent) and Referent Specificity (whether a specific referent is identified), augmented with Number and Countability, while the more specific ones offer numerous semantic types and subtypes, bordering on the idiosyncratic; see work by ~\citet{swan1997english} for an overview. Although none of these variables, individually or in conjunction, can accurately predict article usage, recent work on the classification of a large, manually annotated sample has found that a hierarchical ordering of these same parameters, with Hearer Knowledge at the top, predicts article usage correctly in 93 percent of all cases that allow variation (about $15\%$ of all instances can be considered a set phrase in that only one article can be used, e.g., ``one at a time''~\cite{divjak_romain_milin_2022}.% (Authors, 2022).
%(REF1).

However, deciding whether the interlocutor can be considered as able to identify the referent involves world knowledge, including cultural knowledge; although both Sheffield and Birmingham are home to many universities, when we refer to \emph{the University of Sheffield/Birmingham} we have one particular one in mind, which our interlocutor only knows if they are familiar with the local landscape. In addition, article usage appears to be a matter of what cognitive linguists would call \emph{construal}, or the freedom to present a situation linguistically in different ways. Analysing 3 alternative forced-choice data from 181 native speakers of English who were asked to insert articles that had been removed from longer (200-300 words) texts, \cite{romain_milin_dagmar_2022} relied on Entropy to quantify the restrictiveness of the context and to identify types of contexts in which choice is allowed versus inhibited. They found that some contextual properties, such as Referent Specificity, are rather restrictive, leaving the speaker with little choice in terms of which article to use while other contextual properties, such as Hearer Knowledge, are such that several articles are possible, albeit with slightly different semantic implications In other words, only in situations where the referent is specific do native speakers tend to converge on the same article. 

The English article system thus finds itself in the awkward position of its strongest predictor being open to interpretation. The freedom regarding the interpretation of the top predictor, and the semantic differences it entails, is possibly why second language learners whose first language does not include an article system find the article system notoriously difficult to master. The same can be expected to apply to computational systems who tend to struggle to capture fine-grained meaning nuances, even though they have acquired world knowledge.

\section{Related Work}
\label{section:related-work}
%The most common approach to avoiding the pitfalls of overtly focusing on model performance has been the use of adversarial examples and several such datasets now exist~\citep{sun2020advbert,xu2020elephant,wallace-etal-2019-universal}. Methods for generating adversarial examples exist for both classifiers~\citep{ijcai2018-585} and sequence to sequence models~\citep{cheng2020seq2sick}. Some adversarial datasets consist primarily of particularly difficult examples~\citep{naik-etal-2018-stress} and others focus on a specific phenomenon that models struggle with such as negation~\citep{hossain-etal-2020-analysis}.  

Automatic article prediction has been the focus of study for several decades starting with rule based systems, aimed at improving machine translation~\cite{murata1993determination,bond-etal-1994-countability}. Subsequent machine learning models for article prediction included work by ~\newcite{10.5555/2891730.2891850}, who use decision trees and ~\newcite{han_chodorow_leacock_2006}, who use a maximum entropy classifier to select among a/an, the, or the zero article. 

Article prediction was then clubbed with similar phenomena, such as prepositions and noun numbers, to be included as part of shared tasks on Grammatical Error Correction at CoNLL-2013~\cite{ng-etal-2013-conll} and CoNLL-2014~\cite{ng-etal-2014-conll}. These shared tasks, and their associated datasets, significantly increased interest in article prediction albeit as part of the broader problem of grammatical error correction. More recent methods, such as work by ~\newcite{10.1162/tacl_a_00336}, make use of advances in neural machine translation for grammatical error correction. For an up-to-date and extensive handling of grammatical error correction, including article prediction, we direct readers to the tutorial by ~\newcite{grundkiewicz-etal-2020-crash}. 

Of relevance to the second question we aim to answer, that of how annotator agreement affects model performance, is the work by \newcite{lee-etal-2009-human}, who study the various factors that influence the level of human agreement. Additionally, \citet{ribeiro-etal-2020-beyond} show that state-of-the-art models are better evaluated using a checklist as opposed to traditional metrics, a notion that we supplement in our experimental procedure (Section \ref{section:methodology}). 

\section{Methodology}
\label{section:methodology}
As mentioned in Section \ref{section:intro}, our goal is to understand how language models do on the task of article prediction and how their performance varies with inter-annotator agreement. Our overall methodology for answering these questions involved the following steps: 
%\vspace{-0.4em}
\begin{enumerate}[itemsep=-0.2em]
    \item We start by explicitly adding the null article (Ø) to the British National Corpus (BNC). 
    \item We then set up the task of classifying articles as a token classification (sequence to sequence) task and train a (BERT Base) model. We use 150,000 examples as the training set. 
    \item Using the results of this model, we construct a set of around 2,500 examples, about 30\% of which are selected to be incorrectly tagged by BERT Base. This is to ensure that the evaluation set contains examples from different levels of difficulty. These 2,500 examples are annotated by paid annotators, thus providing us with an evaluation set. 
    \item We compare the performance of human annotators to that of BERT Large, trained on the training set of 150,000 examples from the BNC. 
\end{enumerate}
\vspace{-0.4em}
These results are presented in Section \ref{section:emp-evaluation} along with an analysis. The following sections detail the steps listed above.

\subsection{Data Preparation and Zero article Tagging}
%
% We explicitly add the null article to the BNC using a description provided by language experts, summarised in Table \ref{table:article-types}. The rules used to add this information are included in the program code released as part of this work. 

Table \ref{table:article-types} provides examples of when the zero article is used and we include the scripts used to add zero articles to sentences in the code released with this work. 

\begin{table}[!htbp]
\setlength\dashlinedash{0.2pt}
\setlength\dashlinegap{1.5pt}
\setlength\arrayrulewidth{0.3pt}
{
\small
\def\arraystretch{1.2}
\centering
\begin{tabular}{|L{1.6cm}|L{1.75cm}|L{2.8cm}|}
\hline
 \textbf{Referent Specificity} & \textbf{Noun Count} & \textbf{Example} \\
\hline
\multirow{2}{*}{\makecell{Not Specific, \\ known to \\the hearer}} & Uncountable & \textit{Ø} Pasta is an Italian commodity. \\
& Plural & \textit{Ø} Tigers are magnificent animals.\\
\hdashline
\multirow{2}{*}{\makecell{Not specific, \\ not known \\ to the hearer}} & Uncountable & Can I order Ø rice? \\
 & Plural & I would like Ø better shoes. \\
\hdashline
\multirow{2}{*}{\makecell{Specific, \\ not known \\ to the hearer}} & Uncountable & Ø Soup was served with the meal. \\
& Plural & Ø Engineers were called to the scene. \\
\hline
\end{tabular}
}
\caption{\label{table:article-types} Examples of some occurrences of the \textit{zero article}, also known as the \textit{plural indefinite article}.}
\end{table}

All training and evaluation examples are created to consist of three sentences: the target sentence with one article blanked out and one preceding and one succeeding sentence with no words blanked out. We provide context to ensure that there is sufficient information available to correctly predict an article. Example \ref{example:input}, illustrates one element of the data used. 

{
\small
\begin{exe}
\ex It is a local landmark which received ø national and international recognition and helped turn the tide against the thoughtless demolition of the Sixties. %\\
Still with Booth Shaw, Denison produced \censor{BLANK} radical proposal for ø flats for ø single people in the heart of the city centre. %\\
The site was a rambling and derelict pub, the Royal Hotel, which was originally a Georgian coaching inn. \label{example:input}
\end{exe}
}

\subsection{Model Selection and Training}

Although masked language modelling, which involves ``filling in the blanks'' is most similar to the task at hand, the introduction of the zero article makes this impractical as pre-trained models are not trained on the zero article. Given these limitations we model this as a sequence to sequence task where, as is typical of, the output sequence is required to consist of the token `A', `The' or 'Zero' based on the corresponding article, or the token `O' otherwise. As such, the model makes a prediction associated with \emph{every} input token, not just the one that is masked. 

Based on initial experimentation with different models and hyperparameters (i.e., manual tuning), we settled on the use of BERT fine-tuned on a training set consisting of 150,000 examples for one epoch, based on model performance on a development set (consisting of 30,000 examples). More epochs quickly lead to overfitting. RoBERTa (trained for 6 epochs), despite being considered a more optimised version of BERT, surprisingly does not perform as well as BERT.% (also see Table \ref{table:mcc}) ~\tempred{Todo}. We pick the best of 10 runs with all other hyperparameters set to the default. 

We first use BERT Base, trained on 150,000 examples for 1 epoch, to predict all articles in the target (central) sentence. Based on this initial classification we pick 2,500 examples for manual tagging, such that approximately 30\% of the examples were incorrectly tagged by BERT Base. We perform this additional step to ensure that we pick some examples that are `difficult', as determined by BERT Base's inability to get them right. Finally, BERT Large trained on the same set of examples, is used to predict the articles presented to human annotators. In both cases, we use the models implemented by~\newcite{wolf-etal-2020-transformers}. These results and an analysis are presented in Section \ref{section:emp-evaluation}. Model and hyperparameters are presented in Appendix \ref{app:model}.

\subsection{Human Annotation}

Manual annotation took the format of an online survey modelled after a cloze test. Participants were presented with individual examples consisting of three sentences each, wherein the central sentence had exactly one article omitted and replaced with a blank space, as illustrated in Example \ref{example:input} above. Participants were required to select which article had been omitted from a multiple-choice list that was presented below the sentences.

A total of 2500 sentences were tagged, with each participant tagging 160 randomly selected items. The aim was for each sentence to be tagged by five different participants. Further details on the process including instructions, recruitment, payment and approvals are provided in Appendix \ref{app:annotation}.

\section{Empirical Evaluation and Discussion}
\label{section:emp-evaluation}
The results presented in this section were obtained by evaluating BERT\textsubscript{L} on the same gap filling exercise that was presented to humans. BERT\textsubscript{L} was fine-tuned  5 times on 150,000 training examples and evaluated on a development set which, like the training set, was extracted from the corpus and not human annotated. The training data used consisted of 150,000 examples, of which about 135,000 were ``the'', 60,000 ``a'' and 146,000 ``zero''. The development set consisted of 30,000 examples, of which about 25,000 were ``the'', 12,000 were ```a'' and 25,000 were ``zero''. 

The best performing run on this development set was used for the human annotated test set. Of the 2,500 examples picked for manual annotation, 2,383 were annotated by the required five annotators and this subset was used for evaluation. This evaluation set consists of about 1200 sentences that were annotated by the majority of annotators with ``the'', 500 with ``a'', and about  550 with ``zero''. A further 108 sentences had multiple labels receiving the same number of votes and were thus tied. The complete evaluation set consists of about 150,000 tokens.

\begin{table}[!htbp]
{
\setlength\dashlinedash{0.2pt}
\setlength\dashlinegap{1.5pt}
\setlength\arrayrulewidth{0.3pt}
\footnotesize
\def\arraystretch{1.2}
\centering
\begin{tabular}{|L{0.3cm}|L{2.7cm}| C{0.8cm}|C{0.8cm}|C{0.8cm}|}
\hline
\textbf{} & & \textbf{The} & \textbf{A/An} & \textbf{Zero (Ø)} \\
\hline
%\hdashline
\multirow{2}{*}{\makecell{\rotatebox{90}{All  Data (2384)~~~~}}} & BERT\textsubscript{L} vs 4 Human &  0.580 &	0.659 &	0.589 \\
& BERT\textsubscript{L} vs Corpus &  0.631 & 0.658 &	0.731 \\
& 4 Human vs Corpus &  0.553 & 0.589 & 0.590 \\
& BERT\textsubscript{L} vs Control & 0.488 & 0.573 & 0.514 \\
& 4 Human vs Control & 0.490 & 0.578 & 0.515 \\
& Corpus vs Control & 0.440 & 0.519 & 0.501 \\
\hline
\end{tabular}
\caption{\label{table:mcc-all} Phi coefficient ($\phi$) of correlation between four human annotators (4 Human), BERT Large, a fifth annotator used as a human baseline (Control) and the corpus presented by each article. Number of examples in parenthesis.}
}
\end{table}

Tables \ref{table:mcc-all} and \ref{table:mcc} present the Phi coefficients (Matthews Correlation Coefficient) between four human annotators (4 Human), different models, a fifth human used as a control (Control) and the corpus. Table \ref{table:mcc-all} presents the Phi coefficients across all of the data. Each block in Table \ref{table:mcc} presents Phi correlations between subsets of examples on which either the 4 annotators completely agree (4 agree), exactly three agree (3 agree), or on those examples on which two agreed. In instances other than where all data (Table \ref{table:mcc-all}) is presented, we exclude from our analysis those examples where there is a tie between different articles. Importantly, this results in a different number of examples at each level of agreement presented above (example counts listed in parenthesis). Finally, the last three rows in each block, which provide the correlations with the fifth annotator, provide a baseline or control for comparison.  

Across all data, BERT\textsubscript{L} has a higher correlation with the corpus (BERT\textsubscript{L} vs Corpus) than do the four human annotators (Corpus vs 4 Human) across all articles. While this can be ascribed to the fact that BERT was fine-tuned on a fairly large training set of 150,000 examples, BERT Large also has a higher correlation with the four annotators (BERT\textsubscript{L} vs 4 Human) than they do with the corpus (4 Human vs Corpus) across all but one of the articles on which it misses out by an insignificant margin.% These result are bolstered by the fact that they are substantially higher than the correlations with the fifth annotator used as a control. 

\begin{table}[!htbp]
\setlength\dashlinedash{0.2pt}
\setlength\dashlinegap{1.5pt}
\setlength\arrayrulewidth{0.3pt}
\footnotesize
\def\arraystretch{1.2}
\centering
\begin{tabular}{|L{0.3cm}|L{2.7cm}| C{0.8cm}|C{0.8cm}|C{0.8cm}|}
\hline
\textbf{} & & \textbf{The} & \textbf{A/An} & \textbf{Zero (Ø)} \\
\hline
\multirow{2}{*}{\makecell{\rotatebox{90}{4 Agree (984)~~~~}}} & BERT\textsubscript{L} vs 4 Human & 0.810 & 0.869 & 0.792\\
& BERT\textsubscript{L} vs Corpus & 0.738 & 0.777 & 0.755\\
& 4 Human vs Corpus &  0.787 & 0.822 & 0.767 \\
& BERT\textsubscript{L} vs Control & 0.645 & 0.721 & 0.621\\
& 4 Human vs Control & 0.713 & 0.770 & 0.667 \\
& Corpus vs Control & 0.600 & 0.665 & 0.592 \\
\hdashline
\multirow{2}{*}{\makecell{\rotatebox{90}{3 Agree (886)~~~~}}} & BERT\textsubscript{L} vs 4 Human &  0.545 & 0.617 & 0.626\\
& BERT\textsubscript{L} vs Corpus & 0.605 & 0.639 & 0.719\\
& 4 Human vs Corpus &  0.469 & 0.554 & 0.639 \\
& BERT\textsubscript{L} vs Control & 0.427 & 0.525 & 0.511\\
& 4 Human vs Control & 0.456 & 0.581 & 0.542 \\
& Corpus vs Control & 0.374 & 0.489 & 0.524 \\
\hdashline
\multirow{2}{*}{\makecell{\rotatebox{90}{2 Agree (168)~~~~}}} & BERT\textsubscript{L} vs 4 Human &  0.227 & 0.468 & 0.390 \\
& BERT\textsubscript{L} vs Corpus &  0.501 & 0.549 & 0.692 \\
& 4 Human vs Corpus &  0.280 & 0.344 & 0.403 \\
& BERT\textsubscript{L} vs Control & 0.269 & 0.338 & 0.283\\
& 4 Human vs Control & 0.204 & 0.256 & 0.323\\
& Corpus vs Control & 0.295 & 0.334 & 0.200 \\
\hline
\end{tabular}
\caption{\label{table:mcc} Phi coefficients ($\phi$) at different levels of inter-annotator agreement. See text for details.}
\end{table}

Although BERT has a high correlation with the corpus across all data, a fine-grained analysis based on the possible level of flexibility in article use, as determined by inter-annotator agreement (Table \ref{table:mcc}), shows that this is not always the cast. Surprisingly, when there is least flexibility (i.e. when all four annotators agree) BERT agrees more with human annotators than with the corpus. In fact, in this case (`4 Agree' in Table \ref{table:mcc}) the agreement between BERT and the four annotators is higher than between any other pair. Also interesting is the fact that BERT switches back to being more highly correlated with the corpus when there is any possibility of flexibility (i.e. inter-annotator agreement is not perfect). This is contrary to what we expect as BERT is trained on the corpus and as such we expect to see a higher correlation between BERT and the corpus across all cases. \emph{This behaviour suggest that BERT seems to have access to a high level generalised representation of article use that cannot be ascribed to memory.}

BERT also has a significantly higher correlation with the corpus on the null article than do either the four human annotators or the fifth control annotator except in the case where there is complete agreement between the four annotators (4 Agree). We believe that this is a result of the fact that we insert the null article using a fixed set of rules that deep neural models can easily pick up. Human annotators, on the other hand, seem to find it harder to identify this addition to the article system, except in the more obvious cases.

\section{Conclusions and Future Work}

In this work, we aimed to study the capabilities of pre-trained language models, specifically BERT, on the linguistically relevant task of article prediction that native speakers are intuitively good at but linguists have been unable to formalise adequately, while focusing on how these abilities change with the increased flexibility in article use. Our results show that BERT has a very high correlation with human annotators when there is least flexibility as measured by inter-annotator agreement, but switches to agreeing with the corpus when there is flexibility in article use. These results, we contend, point to BERT having access to a high level generalised representation of article use distinct from memorisation.

We intend to focus future work on better understanding the specifics of this high level representation of article use contained within BERT. Also, the current study is limited in the languages explored and we intend to address this limitation by studying similar intuitive phenomena that evade linguistic description on languages other than English; an example would be aspect in Slavonic languages. Finally, we intend to extend our analysis by comparing BERT's output `confidence' with annotator agreement, similar to methods presented by~\citep{MachineMeetsManEvaluatingthePsychologicalRealityofCorpusbasedProbabilisticModels}.

%%% Dagmar Divjak: For future work we could say that we will look in more detail at the contexts in which raters agree, exploring the hypothesis that those contain nouns that have a specific referent cf hypothesis in Romain et al
% Jan 14, 2022 8:33 AM

% \section{Ethical Considerations}

% Given the critical need to address the climate emergency, experiments were carefully designed and run the minimum number of times required using Amazon Web Services regions reported to be carbon-neutral. We also use the minimum instance capacity when training each of our models, using p2 instances for BERT Base and RoBERTa Base experiments and the higher carbon-footprint P3 instances only for the final BERT Large experiments. 

% While this work focused only on English, we are aware of similar linguistic phenomenon in other languages such as verbal inflection in Polish and hope to work on exploring these phenomenon given time and resources. 

% Human participants were screened to only include native English speakers as this was central to this work. Additionally, to ensure that annotation was not affected by differences in British English and other forms of English, we limited participants to those living in Britain and Ireland. %Finally, we ensured that annotators received a payment.

\section*{Acknowledgements}
We would like to thank Christian Adam, who developed the script for null article tagging, and Daisy Collins for help with setting up and collecting annotations on Qualtrics. 

The manual annotation presented in this work was made possible by the research grant awarded to Harish Tayyar Madabushi by the Paul and Yuanbi Ramsay Research Fund (School of Computer Science, The University of Birmingham). 

This work was also partially supported by the UK EPSRC grant EP/T02450X/1

\bibliography{custom}

\begin{thebibliography}{28}
\expandafter\ifx\csname natexlab\endcsname\relax\def\natexlab#1{#1}\fi

\bibitem[{{BNC Consortium}(2007)}]{bnc2007british}
{BNC Consortium}. 2007.
\newblock British national corpus.
\newblock \emph{Oxford Text Archive Core Collection}.

\bibitem[{Bond et~al.(1994)Bond, Ogura, and
  Ikehara}]{bond-etal-1994-countability}
Francis Bond, Kentaro Ogura, and Satoru Ikehara. 1994.
\newblock \href {https://aclanthology.org/C94-1002} {Countability and number in
  {J}apanese to {E}nglish machine translation}.
\newblock In \emph{{COLING} 1994 Volume 1: The 15th {I}nternational
  {C}onference on {C}omputational {L}inguistics}.

\bibitem[{Chrupa{\l}a and Alishahi(2019)}]{chrupala-alishahi-2019-correlating}
Grzegorz Chrupa{\l}a and Afra Alishahi. 2019.
\newblock \href {https://doi.org/10.18653/v1/P19-1283} {Correlating neural and
  symbolic representations of language}.
\newblock In \emph{Proceedings of the 57th Annual Meeting of the Association
  for Computational Linguistics}, pages 2952--2962, Florence, Italy.
  Association for Computational Linguistics.

\bibitem[{Devlin et~al.(2018)Devlin, Chang, Lee, and
  Toutanova}]{DBLP:journals/corr/abs-1810-04805}
Jacob Devlin, Ming{-}Wei Chang, Kenton Lee, and Kristina Toutanova. 2018.
\newblock \href {http://arxiv.org/abs/1810.04805} {{BERT:} pre-training of deep
  bidirectional transformers for language understanding}.
\newblock \emph{CoRR}, abs/1810.04805.

\bibitem[{Divjak et~al.(2016)Divjak, Dąbrowska, and
  Arppe}]{MachineMeetsManEvaluatingthePsychologicalRealityofCorpusbasedProbabilisticModels}
Dagmar Divjak, Ewa Dąbrowska, and Antti Arppe. 2016.
\newblock \href {https://doi.org/https://doi.org/10.1515/cog-2015-0101}
  {Machine meets man: Evaluating the psychological reality of corpus-based
  probabilistic models}.
\newblock \emph{Cognitive Linguistics}, 27(1):1 -- 33.

\bibitem[{Divjak et~al.(2022)Divjak, Romain, and
  Milin}]{divjak_romain_milin_2022}
Dagmar Divjak, Laurence Romain, and Petar Milin. 2022.
\newblock From their point of view: the article category as a hierarchically
  structured referent tracking system.
\newblock Under revision, Linguistics: an interdisciplinary journal of the
  language sciences.

\bibitem[{Grundkiewicz et~al.(2020)Grundkiewicz, Bryant, and
  Felice}]{grundkiewicz-etal-2020-crash}
Roman Grundkiewicz, Christopher Bryant, and Mariano Felice. 2020.
\newblock \href {https://doi.org/10.18653/v1/2020.coling-tutorials.6} {A crash
  course in automatic grammatical error correction}.
\newblock In \emph{Proceedings of the 28th International Conference on
  Computational Linguistics: Tutorial Abstracts}, pages 33--38, Barcelona,
  Spain (Online). International Committee for Computational Linguistics.

\bibitem[{Han et~al.(2006)Han, Chodorow, and
  LEACOCK}]{han_chodorow_leacock_2006}
NA-RAE Han, MARTIN Chodorow, and CLAUDIA LEACOCK. 2006.
\newblock \href {https://doi.org/10.1017/S1351324906004190} {Detecting errors
  in english article usage by non-native speakers}.
\newblock \emph{Natural Language Engineering}, 12(2):115–129.

\bibitem[{Hewitt and Manning(2019)}]{hewitt-manning-2019-structural}
John Hewitt and Christopher~D. Manning. 2019.
\newblock \href {https://doi.org/10.18653/v1/N19-1419} {{A} structural probe
  for finding syntax in word representations}.
\newblock In \emph{Proceedings of the 2019 Conference of the North {A}merican
  Chapter of the Association for Computational Linguistics: Human Language
  Technologies, Volume 1 (Long and Short Papers)}, pages 4129--4138,
  Minneapolis, Minnesota. Association for Computational Linguistics.

\bibitem[{Huebner(1983)}]{huebner1983longitudinal}
Thomas~G Huebner. 1983.
\newblock \emph{A longitudinal analysis of the acquisition of English by an
  adult Hmong speake}.
\newblock Ph.D. thesis, The University of Hawaii at Mānoa.

\bibitem[{Huebner(1985)}]{https://doi.org/10.1111/j.1467-1770.1985.tb01022.x}
Thorn Huebner. 1985.
\newblock \href
  {https://doi.org/https://doi.org/10.1111/j.1467-1770.1985.tb01022.x} {System
  and variability in interlanguage syntax}.
\newblock \emph{Language Learning}, 35(2):141--163.

\bibitem[{Knight and Chander(1994)}]{10.5555/2891730.2891850}
Kevin Knight and Ishwar Chander. 1994.
\newblock Automated postediting of documents.
\newblock In \emph{Proceedings of the Twelfth AAAI National Conference on
  Artificial Intelligence}, AAAI'94, page 779–784. AAAI Press.

\bibitem[{Lee et~al.(2009)Lee, Tetreault, and Chodorow}]{lee-etal-2009-human}
John Lee, Joel Tetreault, and Martin Chodorow. 2009.
\newblock \href {https://aclanthology.org/W09-3010} {Human evaluation of
  article and noun number usage: Influences of context and construction
  variability}.
\newblock In \emph{Proceedings of the Third Linguistic Annotation Workshop
  ({LAW} {III})}, pages 60--63, Suntec, Singapore. Association for
  Computational Linguistics.

\bibitem[{Lichtarge et~al.(2020)Lichtarge, Alberti, and
  Kumar}]{10.1162/tacl_a_00336}
Jared Lichtarge, Chris Alberti, and Shankar Kumar. 2020.
\newblock \href {https://doi.org/10.1162/tacl_a_00336} {{Data Weighted Training
  Strategies for Grammatical Error Correction}}.
\newblock \emph{Transactions of the Association for Computational Linguistics},
  8:634--646.

\bibitem[{Liu et~al.(2019)Liu, Ott, Goyal, Du, Joshi, Chen, Levy, Lewis,
  Zettlemoyer, and Stoyanov}]{DBLP:journals/corr/abs-1907-11692}
Yinhan Liu, Myle Ott, Naman Goyal, Jingfei Du, Mandar Joshi, Danqi Chen, Omer
  Levy, Mike Lewis, Luke Zettlemoyer, and Veselin Stoyanov. 2019.
\newblock \href {http://arxiv.org/abs/1907.11692} {Roberta: {A} robustly
  optimized {BERT} pretraining approach}.
\newblock \emph{CoRR}, abs/1907.11692.

\bibitem[{Murata(1993)}]{murata1993determination}
M~Murata. 1993.
\newblock Determination of referential property and number of nouns in japanese
  sentences for machine translation into english.
\newblock In \emph{Proc. 5th International Conference on Theoretical and
  Methodological Issues in Machine Translation, Kyoto, Japan, July 1993}, pages
  218--225.

\bibitem[{Ng et~al.(2014)Ng, Wu, Briscoe, Hadiwinoto, Susanto, and
  Bryant}]{ng-etal-2014-conll}
Hwee~Tou Ng, Siew~Mei Wu, Ted Briscoe, Christian Hadiwinoto, Raymond~Hendy
  Susanto, and Christopher Bryant. 2014.
\newblock \href {https://doi.org/10.3115/v1/W14-1701} {The {C}o{NLL}-2014
  shared task on grammatical error correction}.
\newblock In \emph{Proceedings of the Eighteenth Conference on Computational
  Natural Language Learning: Shared Task}, pages 1--14, Baltimore, Maryland.
  Association for Computational Linguistics.

\bibitem[{Ng et~al.(2013)Ng, Wu, Wu, Hadiwinoto, and
  Tetreault}]{ng-etal-2013-conll}
Hwee~Tou Ng, Siew~Mei Wu, Yuanbin Wu, Christian Hadiwinoto, and Joel Tetreault.
  2013.
\newblock \href {https://aclanthology.org/W13-3601} {The {C}o{NLL}-2013 shared
  task on grammatical error correction}.
\newblock In \emph{Proceedings of the Seventeenth Conference on Computational
  Natural Language Learning: Shared Task}, pages 1--12, Sofia, Bulgaria.
  Association for Computational Linguistics.

\bibitem[{Raffel et~al.(2020)Raffel, Shazeer, Roberts, Lee, Narang, Matena,
  Zhou, Li, and Liu}]{JMLR:v21:20-074}
Colin Raffel, Noam Shazeer, Adam Roberts, Katherine Lee, Sharan Narang, Michael
  Matena, Yanqi Zhou, Wei Li, and Peter~J. Liu. 2020.
\newblock \href {http://jmlr.org/papers/v21/20-074.html} {Exploring the limits
  of transfer learning with a unified text-to-text transformer}.
\newblock \emph{Journal of Machine Learning Research}, 21(140):1--67.

\bibitem[{Ribeiro et~al.(2020)Ribeiro, Wu, Guestrin, and
  Singh}]{ribeiro-etal-2020-beyond}
Marco~Tulio Ribeiro, Tongshuang Wu, Carlos Guestrin, and Sameer Singh. 2020.
\newblock \href {https://doi.org/10.18653/v1/2020.acl-main.442} {Beyond
  accuracy: Behavioral testing of {NLP} models with {C}heck{L}ist}.
\newblock In \emph{Proceedings of the 58th Annual Meeting of the Association
  for Computational Linguistics}, pages 4902--4912, Online. Association for
  Computational Linguistics.

\bibitem[{Romain et~al.(2022)Romain, Milin, and
  Dagmar}]{romain_milin_dagmar_2022}
Laurence Romain, Petar Milin, and Dagmar Dagmar. 2022.
\newblock Ruled by construal? framing article choice in english.
\newblock Submitted.

\bibitem[{Swan and Walter(1997)}]{swan1997english}
M.~Swan and C.~Walter. 1997.
\newblock \href {https://books.google.co.uk/books?id=ZUBPvdUx39IC} {\emph{How
  English Works: A Grammar Practice Book ; with Answers}}.
\newblock Oxford English. Oxford University Press.

\bibitem[{Tayyar~Madabushi et~al.(2020)Tayyar~Madabushi, Romain, Divjak, and
  Milin}]{tayyar-madabushi-etal-2020-cxgbert}
Harish Tayyar~Madabushi, Laurence Romain, Dagmar Divjak, and Petar Milin. 2020.
\newblock \href {https://doi.org/10.18653/v1/2020.coling-main.355}
  {{C}x{GBERT}: {BERT} meets construction grammar}.
\newblock In \emph{Proceedings of the 28th International Conference on
  Computational Linguistics}, pages 4020--4032, Barcelona, Spain (Online).
  International Committee on Computational Linguistics.

\bibitem[{Tenney et~al.(2019{\natexlab{a}})Tenney, Das, and
  Pavlick}]{DBLP:journals/corr/abs-1905-05950}
Ian Tenney, Dipanjan Das, and Ellie Pavlick. 2019{\natexlab{a}}.
\newblock \href {http://arxiv.org/abs/1905.05950} {{BERT} rediscovers the
  classical {NLP} pipeline}.
\newblock \emph{CoRR}, abs/1905.05950.

\bibitem[{Tenney et~al.(2019{\natexlab{b}})Tenney, Xia, Chen, Wang, Poliak,
  McCoy, Kim, Durme, Bowman, Das, and Pavlick}]{tenney2018what}
Ian Tenney, Patrick Xia, Berlin Chen, Alex Wang, Adam Poliak, R~Thomas McCoy,
  Najoung Kim, Benjamin~Van Durme, Sam Bowman, Dipanjan Das, and Ellie Pavlick.
  2019{\natexlab{b}}.
\newblock \href {https://openreview.net/forum?id=SJzSgnRcKX} {What do you learn
  from context? probing for sentence structure in contextualized word
  representations}.
\newblock In \emph{International Conference on Learning Representations}.

\bibitem[{Thomas(1989)}]{thomas_1989}
Margaret Thomas. 1989.
\newblock \href {https://doi.org/10.1017/S0142716400008663} {The acquisition of
  english articles by first- and second-language learners}.
\newblock \emph{Applied Psycholinguistics}, 10(3):335–355.

\bibitem[{Wolf et~al.(2020)Wolf, Debut, Sanh, Chaumond, Delangue, Moi, Cistac,
  Rault, Louf, Funtowicz, Davison, Shleifer, von Platen, Ma, Jernite, Plu, Xu,
  Scao, Gugger, Drame, Lhoest, and Rush}]{wolf-etal-2020-transformers}
Thomas Wolf, Lysandre Debut, Victor Sanh, Julien Chaumond, Clement Delangue,
  Anthony Moi, Pierric Cistac, Tim Rault, Rémi Louf, Morgan Funtowicz, Joe
  Davison, Sam Shleifer, Patrick von Platen, Clara Ma, Yacine Jernite, Julien
  Plu, Canwen Xu, Teven~Le Scao, Sylvain Gugger, Mariama Drame, Quentin Lhoest,
  and Alexander~M. Rush. 2020.
\newblock \href {https://www.aclweb.org/anthology/2020.emnlp-demos.6}
  {Transformers: State-of-the-art natural language processing}.
\newblock In \emph{Proceedings of the 2020 Conference on Empirical Methods in
  Natural Language Processing: System Demonstrations}, pages 38--45, Online.
  Association for Computational Linguistics.

\bibitem[{Zhang et~al.(2020)Zhang, Yang, and Zhao}]{zhang2020retrospective}
Zhuosheng Zhang, Junjie Yang, and Hai Zhao. 2020.
\newblock \href {http://arxiv.org/abs/2001.09694} {Retrospective reader for
  machine reading comprehension}.

\end{thebibliography}
\bibliographystyle{acl_natbib}

\appendix

\section{The BNC}
\label{app:bnc-licence}
The dataset used in the experiments presented in this work is extracted from the British National Corpus (BNC) distributed by the University of Oxford on behalf of the BNC Consortium and is consistent with its intended use. We extract sentences from both the spoken (BNC 2014 release) and the written (BNC 1994 release) versions of the BNC. Examples cited within the paper have been extracted from the BNC and all rights in the texts cited are reserved. We make use of the BNC to ensure that we use a well balanced data source that does not uniquely identify individuals or include offensive content. Detailed statistics pertaining to the BNC are available on the BNC website\footnote{\url{http://www.natcorp.ox.ac.uk/corpus/index.xml?ID=intro}}. 

The BNC is available under the BNC User Licence\footnote{\url{http://www.natcorp.ox.ac.uk/docs/licence.html}} and given that we build our dataset from the BNC, access to our dataset is subject to access to the BNC. 

\section{Model, Training, Hyperparameter and Hardware Details}
\label{app:model}

%The training data used consisted of 150,000 examples consisting of about 8.3M tokens, of which about 135,000 were ``the'', 60,000 ``a'' and 146,000 ``zero''. The development set consisted of 30,000 examples with about 1.64M tokens, of which about 25,000 were ``the'', 12,000 were ```a'' and 25,000 were ``zero''. The evaluation set used to obtain the results consisted of 2,384 examples with about 150,000 tokens, of which about 1000 where ``the'', 500 ``a'', and 750 ``zero''. 

For our experiments, we make use of BERT Base, which consists of 110 million parameters and BERT Large consisting of 340 million parameters. We use the default hyperparameters for both models except in changing the number of epochs to 1 and the maximum input sequence length to 150. This was based on our initial experimentation wherein  we found that more epochs quickly lead to overfitting. In particular, we run our experiments using the Hugging Face Transformers implementation available online\footnote{\url{https://github.com/huggingface/transformers/blob/master/examples/legacy/token-classification/run_ner.py}}. 
%RoBERTa, we found, (trained for 6 epochs), surprisingly does not perform as well as BERT (also see Table \ref{table:mcc}) ~\tempred{Todo}. 

Models were trained using a Tesla V 100 GPU, and the entire training and optimisation process took approximately forty hours. 

Models were run multiple times, each with a different random seed so as to avoid local minimum. In each case, models were evaluated on the development set which, like the training set was extracted from the corpus and not manually annotated. The best performing model on the development set was used for subsequent experiments. The results over 10 different random seeds on the development set for BERT Base are presented in Table \ref{table:dev-randomseed}.
\begin{table}[!htbp]
\setlength\dashlinedash{0.2pt}
\setlength\dashlinegap{1.5pt}
\setlength\arrayrulewidth{0.3pt}
\small
\def\arraystretch{1.2}
\centering
\begin{tabular}{|L{1cm}|C{2cm}|}
\hline
Run No. & Dev F1 \\
\hline
 1 & 0.8940 \\
 2 & 0.8936 \\
 3 & 0.8953 \\
 4 & 0.8942 \\
 5 & 0.8957 \\
 6 & 0.8930 \\
 7 & 0.8941 \\
 8 & 0.8947 \\
 9 & 0.8936 \\
 10 & 0.8944 \\
\hline
\end{tabular}
\caption{\label{table:dev-randomseed}Results over 10 different random seeds on the development set for BERT Base -- used to pick the best run used in subsequent experiments. We note that the variation in results across radom seeds isn't significant due to the large training set used.}
\end{table}. 

We calculate the Phi coefficients ($\phi$) in R (version 4.0.3) using the psych package (version 2.0.9). 

\section{Annotation Details}
\label{app:annotation}
The annotation was done using Qualtrics and participants were recruited through Prolific. Each participant was compensated £3.75 for annotating approximately 160 examples, which took participants an average of 42 minutes, a little over the 30 minutes we estimated it would take. We recruited a total of 108 annotators of whom 68 were female and 40 were male. Most annotators had a Bachelor's degree or had attended some college, and close to 65\% of them were between the ages of 20 and 40. 

Participants, who were all native speakers of British English and residing in the UK or Ireland (due to the use of the BNC), were instructed to read all three sentences before choosing which article they would fill the gap with. Four quality control questions were included in order to make sure that participants were paying attention. 

The exact quality control questions were chosen following a pilot study run on 15 participants - a manual analysis of these results by linguists indicated that those who failed to correctly answer any one of these quality control questions, considered to be relatively straightforward, seemed to do little better than chance overall. If any one of the quality control questions were answered incorrectly, participants were not allowed to continue with the survey.

The risks associated with annotation are two fold: The first is to do with the risk of annotators not being representative of the general population. As such, we placed no restrictions on the demographics of our annotators except as required by the study. That is, we recruited fluent English speakers from the UK and Ireland, to ensure that they speak British English, consistent with our use of the BNC. The second risk is to do with annotators not being treated fairly. To ensure that this was not the case, we paid annotators a sum of £3.75 for what we estimated, based on our internal trials, would constitute 30 minutes of work. In addition, data collection was run with the approval of the ethics committee at the University.

\subsection{Instructions to Annotators}

Thank you for agreeing to take part in this study. For participating in the study you will earn £3.75. This study is run with the approval of the ethics committee at the University.

 If you have any questions about the survey please contact me, Dr Harish Tayyar Madabushi at: H.TayyarMadabushi.1@bham.ac.uk.
\newline
\newline
\noindent\textbf{Instructions}

\noindent Please read these instructions carefully before continuing to fill in this survey. 

In this study you will be presented with three sentences on each trial. In the middle sentence, one word is missing and it is your task to provide it; it can be either a(n), the or ZERO. In the first and last sentence, all words are provided. Please read all three sentences before filling the gap.
\newline
\newline
\noindent\textbf{Example}

\noindent Consider the following example where the special character `Ø' represents locations where an article could have occurred, but, in this particular case, does not: 

But there is no escape for Ø non - runners , who are required to sign up for Ø light duties.
That takes \censor{BLANK} care of Sunday .
We cannot refuse, because we are in Ø awe of the formidable women running the PTA.

You are required to fill in the \censor{BLANK} with one of:

\begin{enumerate}
    \item a/an
    \item the
    \item Zero (Ø) 
\end{enumerate}

In the example above, the correct answer is Zero (Ø). 
\newline
\newline
\noindent\textbf{Instructions}

\noindent This survey consists of approximately 170 questions and should take you about 30 minutes to complete. 

IMPORTANT: Some of these questions - the quality check questions - will be used to perform a quality check and will be presented at random points in this survey. If you get too many of the quality check questions incorrect, your submission may be rejected. Please pay attention to the answers you provide as rejected submissions are not eligible for payment.

Thank you very much for taking the time to participate in this study. You will first need to answer some questions about your background, followed by a few benchmark questions, before you start on the bulk of the survey.

\end{document}